\documentclass[11pt,onecolumn,journal,draftcls]{IEEEtran}
\pdfoutput=1 

\usepackage{cite}

\usepackage[utf8]{inputenc} 
\usepackage[T1]{fontenc}    
\usepackage{url}            
\usepackage{booktabs}       
\usepackage{amsfonts}       
\usepackage{nicefrac}       
\usepackage{microtype}      

\usepackage{amssymb,amsmath}
\usepackage{bbm}
\usepackage{color}
\usepackage{xcolor}
\usepackage{graphicx}
\usepackage{subfigure}

\usepackage{pgfplots}
\pgfplotsset{compat=newest}

\usepackage{tikz}
\usepackage{textcomp}
\usetikzlibrary{plotmarks}
\usetikzlibrary{shapes,arrows}
\usetikzlibrary{backgrounds,calc,positioning}

\usepackage{amsthm}
\usepackage{centernot}

\usepackage{juremath}

\title{Learning to Succeed while Teaching to Fail:\\ Privacy in Closed Machine Learning Systems}

\author{Jure~Sokoli{\'c},
		Qiang~Qiu,
		Miguel~R.~D.~Rodrigues,
		and~Guillermo~Sapiro
		\thanks{J. Sokoli{\'c} and M. R. D. Rodrigues are with the Department of Electronic and Electrical Engineering, Univeristy College London, London, UK (e-mail:\texttt{ \{jure.sokolic.13, m.rodrigues\}@ucl.ac.uk}).}	
		\thanks{Q. Qiu and G. Sapiro are with the Department of Electrical and Computer Engineering, Duke University, NC, USA (e-mail: \texttt{ \{qiang.qiu, guillermo.sapiro\}@duke.edu}).}		
\thanks{The work of Guillermo Sapiro was partially supported by NSF, ONR, ARO, NGA. }
}

\begin{document}

\maketitle

\begin{abstract}
Security, privacy, and fairness have become critical in the era of data science and machine learning. More and more we see that achieving universally secure, private, and fair systems is practically impossible. We have seen for example how generative adversarial networks can be used to learn about the expected private training data; how the exploitation of additional data can reveal private information in the original one; and how what looks like unrelated features can teach us about each other. Confronted with this challenge, in this paper we open a new line of research, where the security, privacy, and fairness is learned and used in a closed environment. The goal is to ensure that a given entity (e.g., the company or the government), trusted to infer certain information with our data, is blocked from inferring protected information from it. For example, a hospital might be allowed to produce diagnosis on the patient (the positive task), without being able to infer the gender of the subject (negative task). Similarly, a company can guarantee that internally it is not using the provided data for any undesired task, an important goal that is not contradicting the virtually impossible challenge of blocking everybody from the undesired task. We design a system that learns to succeed on the positive task while simultaneously fail at the negative one, and illustrate this with challenging cases where the positive task is actually harder than the negative one being blocked. Fairness, to the information in the negative task, is often automatically obtained as a result of this proposed approach. The particular framework and examples open the door to security, privacy, and fairness in very important closed scenarios, ranging from private data accumulation companies like social networks to law-enforcement and hospitals.
\end{abstract}

\section{Introduction}

Advances in machine learning and data science allow to develop product and services that leverage individuals data to provide valuable services. At the same time care must be taken to provide an appropriate amount of protection for users in order to adhere to various legal and ethical constraints. 

In this work we envision the following scenario: An entity (e.g., the company or the government) wants to offer a service to its users based on their data. However, some users want to be able to prevent the entity from inferring certain sensitive information from their data. The need for this may come from security, fairness, or privacy concerns. For example, a hospital might be allowed to produce diagnosis on the patient, without being able to infer some other irrelevant condition of the subject. We may allow identity verification for security purposes, but we want to be sure that we can not be discriminated based on gender, race, or age, guaranteeing that such information can not even be inferred from the provided data. This motivates the main question that we try to answer in this work:

\vspace{0.2cm}
\textit{How can we design a system that can provide a valuable service to a user and offer protection to their sensitive data at the same time?}
\vspace{0.2cm}

The benefits of such a system are twofold. First, the users will be more comfortable because their privacy is respected for the tasks the user wants to block. Second, the system will offer the entity offering the service a principled approach to prove that they are not using/inferring users' sensitive information in case of legal/ethical disputes. 

Before describing the proposed framework we need to further address two issues that motivate our proposed system. First, can we design a system that is universal, meaning capable to prevent the undesired task by any other system? And second, is the problem that we are trying to solve trivial? The negative answers to these questions are important motivations to the framework here introduced.


Regarding the first question, there  are many examples in practice where additional sources of information lead to attacks on the systems that were designed to be private. For example, the authors in \cite{Narayanan2008} showed how to break the anonymity of the publicly released Netflix Prize Dataset. They achieved that by using other public datasets and an appropriate algorithm. It has been shown in \cite{Oh2016} that  a person in an image can be identified based on the social network graph and the photos of other users even when the faces in the images are obfuscated. The source and power of this additional data may be very hard to anticipate. Another example is the unexpected possibility to infer cardio-vascular health, a task we might want to protect, from regular credit scores as those commonly provided to get mortgage authorization \cite{Israel2014}. 

The impossibility of universal privacy protection just exemplified has also been studied extensively in the domain of differential privacy \cite{Dwork2008}, where a number of authors have shown that assumptions about the data or the adversary must be made in order to be able to provide utility  \cite{Dwork2010, Kifer2011a,  Kifer2014}. Similarly, in the fairness domain the authors in \cite{Hardt2016} propose a method to remove discrimination from any learned predictor and note that their fairness objective is not universal, but rather domain specific. As a consequence their framework does not aim to certify fairness but rather provide a set of tools which can be used to improve the fairness of a system.

On the other hand, one may argue that users can prevent inference of sensitive variables by simply sharing less data. For example, \cite{AliOssia} shows that if the feature vector representing a face image is sufficiently low dimensional, one can infer the gender of the person, but not his/her identity. However, one may also be interested in face verification, which is not possible if the features do not provide sufficient information, and be agnostic to gender at the same time. This task is much harder, meaning the desired task (verification) is harder than the protected task (gender detection), and the appropriate assumptions must be made, as we will be shown in the sequel.

In view of the issues presented above we propose a novel framework that rather than being universal, assumes a closed environment where privacy with respect to particular sensitive variables is desired. The goal is to ensure that a given entity, trusted to infer certain information with our data, is blocked from inferring sensitive information from it. For example, a hospital might be allowed to produce diagnosis on the patient (the positive task), without being able to infer the irrelevant gender of the subject (negative task). Similarly, a company can guarantee they internally are not using the provided data for any undesired task, an important goal that is not contradicting the virtually impossible challenge of blocking everybody from the undesired task. 

\begin{figure}[t]
\centering
	\includegraphics[height=0.3\linewidth]{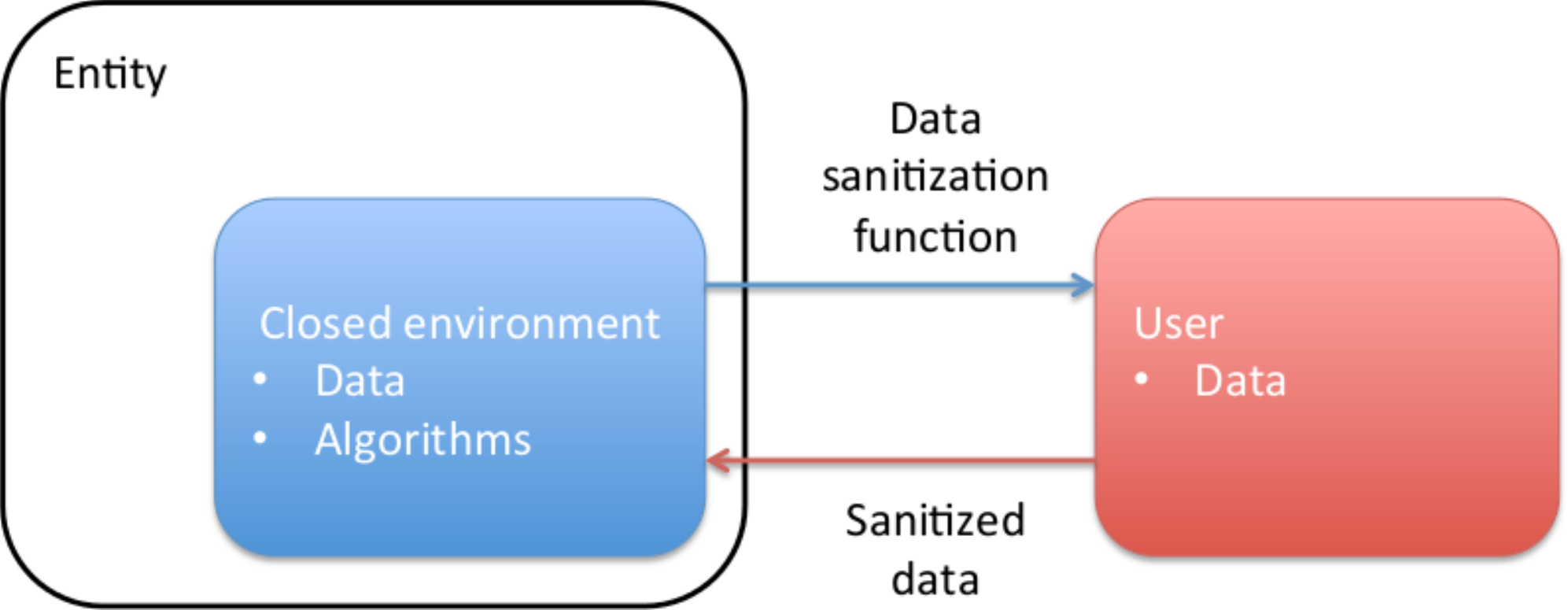}	
	\caption{The entity provides a service to its users in a closed environment with a set of data and a set of algorithms. Given a sensitive variable that may be infered from raw user data, the entity designs a \textit{data sanitization function}, which transformes users' data in such a way that the sensitive variables may not be infered whithin the closed environment. The data sanitization function is then either shared with the user who can now share \textit{sanitized data} or it is applied to the data before the users' data enters the entity.} \label{fig:idea}

\end{figure}

To achieve this, we assume an environment where the entity declares a set of (training) data and a set of tools (machine learning algorithms) that will be used to provide a certain service (positive task) to users. We then design a data sanitization function whereby users provide data to the entity that prevents inference of sensitive variables (negative tasks) within the constrained environment. This general idea is visualized in Figure~\ref{fig:idea}.


The proposed framework is describe in detail in Section~\ref{sec:framework}. A critical application is presented in Section~\ref{sec:applications}. Related literature is reviewed and discussed in Section~\ref{sec:related_work}. Finally, the paper is concluded in Section~\ref{sec:conclusion}. Details on the experimental procedures are provided in the Appendix.

\section{Proposed Framework} \label{sec:framework}

Our framework consists of a \textit{closed environment} with a set of data $\sZ$ and a set of algorithms $\sA$. In particular, we assume that the training set consist of $m$ training examples: $\sZ = \{ z_i \}^m_{i=1}$. An element of $\sZ$ is a triplet $z = (x, t, s)$, where $x \in \sX$ represents a feature vector is some feature space $\sX$, $t$ is a label associated with utility (positive task) and $s$ is a label associated with sensitive data (negative task). For the sake of simplicity we assume that $t$ and $s$ are binary, i.e., $t,s \in \{-1,1\}$. The set of algorithms $\sA$ consists of algorithms $A$ that map the training set to a predictor
\begin{IEEEeqnarray}{rCl}
	A: \sZ \to q_{A(\sZ)} \quad  A \in \sA \,,
\end{IEEEeqnarray}
where the predictor is of the form
\begin{IEEEeqnarray}{rCl}
	q_{A(\sZ)}: \sX \to \{-1,1 \}\,.
\end{IEEEeqnarray}
The training data and the algorithms are used to provide a service to the users, where the users' data  consists of feature vectors $x$ and it is denoted by $\sU = \{ (x)_i \}_{i=1}^n$. In particular, the entity provides a service to the user via the prediction of the utility variable $t$. However, since the feature vector may also be used to predict the sensitive variable $s$, care must be taken to provide protection against that. As discussed before, on one hand, a user might not use the service because of the concerns of revealing her sensitive information. On the other hand, the entity wants to be able to prove that they are not misusing users data by predicting the sensitive variable $s$ against the user's preference, or for example due to fairness. We assume that the training set has no privacy restrictions.

As discussed in the Introduction, giving guarantees about non-identifiability of certain variables in the ``open world'' where new data or new algorithms are readily available is very likely impossible. Therefore, what we want to do is to design mechanisms that will prevent prediction of sensitive variables within this important closed environment. Such methods are very relevant in practice: i) they create an additional level of trust between the users and the entity; ii) it gives the entity an elegant way to certify that sensitive information is not being abused within their closed environment; and iii) can provide additional value such as fairness.\footnote{Such a closed environment can be achieved for example by specifying what algorithm has access to what data in a manner similar to the implementation of human access control in digital environments.}

We require the entity to define a utility metric, a privacy metric, and to  design a data sanitization function that will achieve privacy while providing utility.  Of particular challenge is the case when inferring $s$ (private data) is actually easier than inferring the actual authorized information $t$. This is here addressed for the first time.

\paragraph{Utility Metric, Privacy Metric and Data Sanitization.}


First, we will assume that the entity has a test set $\sZ'$ with $m'$ samples where the samples are drawn from the same distribution as the training examples. We define both the utility metric and the privacy metric based on the testing set. This is needed since a representative test set is exploited to gain confidence about the utility of the machine learning model. The same test can be used to verify that given the training set $\sZ$ and the set of algorithms $\sA$, no meaningful conclusion about the sensitive variable $s$ can be drawn from the data.

The entity will then design a data sanitization function 
\begin{IEEEeqnarray}{rCl}
	f: \sX \to \sX \,,
\end{IEEEeqnarray}
such that both utility and privacy will be satisfied. Note that $f$ maps to the same feature space $\sX$ so that sanitized data $f(x)$ may be used as the input to a predictor $q_{A(\sZ)}$. We will evaluate the data sanitization function on the sanitized test set which will be denoted as $\sZ'_f$:
\begin{IEEEeqnarray}{rCl}
	\sZ'_f = \{(f(x), s,t )) : (x,s,t) \in \sZ' \}.
\end{IEEEeqnarray}
In particular, the utility metric of the closed system  will be denoted by 
\begin{IEEEeqnarray}{rCl}
	u(\sA, \sZ, \sZ'_f) \in [0,1] \,,
\end{IEEEeqnarray}
and the privacy metric will be denoted as
\begin{IEEEeqnarray}{rCl}
	p(\sA, \sZ, \sZ'_f) \in [0,1] \,.
\end{IEEEeqnarray}
The goal is to design such $f$ that both utility and privacy metrics are as close to 1 as possible.

For example, denote by $\text{CA}_t(A, \sZ, \sZ_f')$ the classification accuracy of the utility task of the predictor $q_{A(\sZ)}$ trained on the training set $\sZ$ and evaluated on the sanitized test set $\sZ_f'$:
\begin{IEEEeqnarray}{rCl}
	\text{CA}_t(A, \sZ, \sZ_f') = \frac{1}{|\sZ'_f|} \sum_{(f(x),s,t) \in \sZ'_f }  \left(1-\ell(q_{A(\sZ)}(f(x)), t) \right)\,,
\end{IEEEeqnarray}
where $\ell$ is a 0-1 loss. Then we might define the utility metric as the classification accuracy of the best possible predictor given the training set $\sZ$ and the set of algorithms $\sA$:
\begin{IEEEeqnarray}{rCl}
	u(\sA, \sZ, \sZ'_f) = \max_{A \in \sA} \text{CA}(A, \sZ, \sZ_f') \,.
\end{IEEEeqnarray}
Similarly, denote by $\text{CA}_s(A, \sZ, \sZ_f')$ the classification accuracy of the sensitive task of the predictor $q_{A(\sZ)}$ trained on the training set $\sZ$ and evaluated on the sanitized test set $\sZ_f'$:
\begin{IEEEeqnarray}{rCl}
	\text{CA}_s(A, \sZ, \sZ_f') = \frac{1}{|\sZ'_f|} \sum_{(f(x),s,t) \in \sZ'_f }  \left(1-\ell(q_{A(\sZ)}(f(x)), s) \right)\,.
\end{IEEEeqnarray}
Assume also that the prior probabilities of the sensitive labels are balanced, $p(s=1) = p(s=0) = 0.5$. Then the privacy metric can be defined as the deviation of the the most accurate predictor given the training set $\sZ$ and the set of algorithms $\sA$ from the ``random'' classifier:
\begin{IEEEeqnarray}{rCl}
	p(\sA, \sZ, \sZ'_f) = \min_{A \in \sA} 1 - 2\cdot| \text{CA}_s(A, \sZ, \sZ_f') -0.5| \,.
\end{IEEEeqnarray}
Note that the definitions of privacy and the utility are application specific. For example, we may define the utility or privacy metrics via averages over different users. However, we could use more restrictive definitions where utility or privacy are defined per user. The novel framework here introduced is independent of the particular selection. A concrete example is presented in Section~\ref{sec:applications}.

The entity might now collect users data via the data sanitization function $f$ into the environment with training data $\sZ$ and with the set of algorithms $\sA$. The closed environment is certified with the utility $u(\sA, \sZ, \sZ'_f)$ and the privacy $p(\sA, \sZ, \sZ'_f)$, which may be communicated to the user or a possible regulatory entity. 

We have presented a framework to protect users privacy based explicitly on the predefined data and algorithms. The guarantees do not extend to algorithms outside $\sA$ and additional data (side information) outside $\sZ$. Although these properties would be desirable, as previously described a large body of literature suggests that universal privacy protection is only possible at the expense of utility. 

\section{Sample Applications} \label{sec:applications}

We now demonstrate the proposed framework with an application involving face images and different types of closed environments.

\subsection{Face verification and gender recognition from face images}

 We will focus on the tasks of face verification\footnote{Note that face verification verifies if two images belong belong to the same person (consider for example the ID verification at the airport or the entrance to a facility). It does not identify a person and therefore does not reveal gender.} and gender classification. In particular, we will consider the face verification as the utility task and the gender recognition as the sensitive task. This may be useful in various applications where we want to re-identify users, but want to prevent gender discrimination/information within the system. This of course goes beyond fairness, where the goal is only to make sure that the system's performance is independent of gender, but not to block the gender detection. 

Note that for the opposite case, where gender classification is the utility task and face verification is the sensitive task is also relevant. However, in this case the undesired task can be easily blocked by simply using a sufficiently low dimensional projection, as shown in \cite{AliOssia}, and is therefore less challenging.

\subsubsection{Dataset and Tools}

\paragraph{Dataset.} We use the FaceScrub dataset \cite{Ng2014}, containing approximately 100,000 face images of 530 individuals, with 265 males and 265 females. For each image $x$ is obtained by using the features of the last layer of the VGG-face network \cite{Parkhi2015a} projected to dimension 1000 using principal component analysis. Associated with each feature is the identity label $t$ and the gender label $s$. Note that our positive task is verification and not identification, therefore, given two examples $(x,t,s)$ and $(x',t',s')$ we want to establish either $t=t'$ or $t\neq t'$ based on the features $x$ and $x'$. The negative task is gender classification where the features $x$ are used to predict the gender $s$.

We create three subsets from the original dataset:
\begin{itemize}
	\item Training set $\sZ = \{(x,t,s)_i\}_{i=1}^m$, which represents data in the closed environment and is used to design the data sanitization function;
	\item Test set $\sZ' = \{(x,t,s)_i\}_{i=1}^{m'}$, which is used to evaluate the utility and privacy metrics. The sanitized version of the test set is denoted as $\sZ_f' = \{(f(x),t,s)_i\}_{i=1}^{m'}$;
	\item User set $\sU = \{ (x)_i \}_{i=1}^n$, which represents users' data.
\end{itemize}
The three sets $\sZ$, $\sZ'$ and $\sU$ have a balanced ratio of males and females and do not share identities.

\paragraph{Metrics.} We use the Area Under the Curve (AUC) of the Receiver Operating Characteristic (ROC) curve to measure success of the face verification task and the gender classification accuracy (CA) to measure success of the gender classification task. Therefore, the utility and the privacy metrics for this particular application are defined as 
\begin{IEEEeqnarray}{rCl}
 	u(\sA, \sZ, \sZ'_f) = \text{AUC}(\sZ'_f) \,
\end{IEEEeqnarray}
and
\begin{IEEEeqnarray}{rCl}
	p(\sA, \sZ, \sZ'_f) = \min_{A \in \sA} 1 - 2\cdot| \text{CA}_s(A, \sZ, \sZ_f') -0.5| \,,
\end{IEEEeqnarray}
respectively.
\paragraph{Algorithms.}

We test using 3 algorithms, helping us to present the proposed framework and not just an instance of it:
\begin{itemize}
	\item Verification algorithm (VA): it takes two feature vectors and it outputs the cosine similarity between the two vectors;
	
	\item Support Vector Machine (SVM) classifier;
	
	\item Random Forest (RF) classifier.
\end{itemize}
The exact training procedure and hyper-parameter settings are described in the Appendix.

\paragraph{Data Sanitization.}

We consider two data sanitization techniques that are fully described in the Appendix:
\begin{itemize}
	\item Data Sanitization with Linear Projection:
	\begin{IEEEeqnarray}{rCl}
		f(x) = P x \,,
	\end{IEEEeqnarray}
	where $P$ is a linear projection matrix. 
	
	\item Data Sanitization with Maximum Mean Discrepancy Transformation:
	\begin{IEEEeqnarray}{rCl}
		f(x) = x + n(x) \,,
	\end{IEEEeqnarray}
	where $n(x)$ is a ``noise'' vector that depends non-linearly on $x$.
\end{itemize}
The data sanitization function are designed so that they maximize the privacy and minimally affect the utility.

\subsubsection{Results}

We now explore various scenarios where we prevent gender classification without compromising verification within a closed environment. In all the cases we will assume that the training set is part of the closed environment. We will then consider the different examples of data sanitization functions and the different algorithms allowed in the closed environment. 

\paragraph{Baseline.} 

First, we consider a baseline scenario, where the closed environment uses the training set $\sZ$ and all three algorithms: $\sA = \{\text{VA},\text{SVM}, \text{RF} \}$ and does not do any data sanitization. We report the values of the utility metric and the privacy metric as measured on the test set $\sZ'$. We also report the gender classification accuracy of the SVM classifier, the gender classification accuracy of the RF classifier and the verification AUC on the user data. The results are reported in Table~\ref{tab:baseline}.

Note that the utility metric is close to 1 and the privacy metric is close to zero. This is expected as no data sanitization is used. The gender classification accuracies on the data in the user set are high, which implies low privacy, and the verification AUC on the user data is also high, which implies high utility. 

\begin{table}[t]
\centering
\caption{Baseline. Setup: $f(x) = x$, $\sZ$, $\sA = \{\text{VA},\text{SVM}, \text{RF} \}$.} \label{tab:baseline}
\begin{tabular}{ccccc}
\toprule
 & & \multicolumn{3}{c}{User data} \\
\cmidrule{3-5} 
Utility metric & Privacy metric & Gender acc. (SVM) & Gender acc. (RF) & Verification AUC\\
\midrule
0.9607 & 0.0672 & 96.73 \% & 95.37 \% & 96.10 \% \\
\bottomrule
\end{tabular} 
\vspace{0.5cm}
\centering
\caption{Case 1. Setup: $f(x) = Px$, $\sZ$, $\sA = \{\text{VA},\text{SVM} \}$.} \label{tab:linear}
\begin{tabular}{ccccc}
\toprule
 & & \multicolumn{3}{c}{User data} \\
\cmidrule{3-5} 
Utility metric & Privacy metric & Gender acc. (SVM) & Gender acc. (RF) & Verification AUC\\
\midrule
0.9600 & 0.8892 & 45.81 \% & / & 96.08 \% \\ 
\bottomrule
\end{tabular} 
\vspace{0.5cm}
\centering
\caption{Case 2.  Setup: $f(x) = x + n(x)$, $\sZ$, $\sA = \{\text{VA},\text{SVM}, \text{RF} \}$.} \label{tab:mmd}
\begin{tabular}{ccccc}
\toprule
 & & \multicolumn{3}{c}{User data} \\
\cmidrule{3-5} 
Utility metric & Privacy metric & Gender acc. (SVM) & Gender acc. (RF) & Verification AUC\\
\midrule
0.9493 & 0.8713 & 50.27 \% & 53.84 \% & 95.07 \% \\
\bottomrule
\end{tabular} 
\end{table}

\paragraph{Case 1: Linear sanitization and SVMs.}

Here we set the closed environment to include the training set $\sZ$, algorithms: $\sA = \{\text{VA},\text{SVM} \}$, and the linear sanitization function for the users' data. The values of the utility metric and the privacy metric as measured on the test set $\sZ'_f$ are reported in Table~\ref{tab:linear}. The gender classification accuracy of the SVM classifier and the verification AUC on the user data are reported as well. Note that the the gender classification accuracy of the RF classifier is not reported because RF classifier is not in the closed environment.

The utility metric in this case if very close to the baseline. Therefore, the utility is preserved by the data sanitization function. The privacy metric, on the other hand, is much higher than in the baseline example, which means that gender recognition does not perform well on the users' data. These observations are also supported by a low gender classification accuracy of the SVM classifier and the verification AUC on the user dataset, which dropped for only $0.02\%$ compared to the baseline. Therefore, we have simultaneously achieved privacy  (and gender fairness as a consequence) and utility within the specified environment.

\paragraph{Case 2: MMD sanitization, SVMs and RFs.} Now we extend the closed environment from case 1 to include the stronger RF classifiers: $\sA = \{\text{VA},\text{SVM}, \text{RF} \}$. The linear sanitization function is not effective when used with the RF classifiers, therefore we use the MMD data sanitization. The results following the format of the baseline and case 1 are presented in Table~\ref{tab:mmd}.

Note that the utility metric and the privacy metric are slightly lower than in the case 1. Nevertheless, the verification AUC on the user dataset dropped by $1.03\%$ compared to the baseline, whereas the SVM and RF gender classifiers trained in the closed environment achieved accuracies $50.27 \%$ and $53.84\%$, respectively. We can claim that the data sanitization is therefore effective.



\section{Related Work} \label{sec:related_work}


While the concept of closed environment privacy as here defined is new, and the formulation presented in the previous section should be considered as one possible realization of it, it is related to other contributions in the literature. We give a review of some of these works next.

\paragraph{Privacy.}

The privacy problem has been widely studied in the data mining and database communities, where differential privacy \cite{Dwork2008} is one of the most popular approaches.

Differentially private data release mechanisms ensure that an adversary may not establish a presence or absence of an individual in a database \cite{Wasserman2010}. An important property, emphasized in the differential privacy literature, is the fact that there is no need for assumptions about the side information of an adversary. However, such universality comes at an expense of utility, as shown in \cite{Dwork2010,Kifer2011a,Kifer2014}. Our proposed framework differs from the differential privacy framework by assuming a closed environment where the data and tools are predefined/pre-trained. A consequence is that we can maintain utility, while still offering privacy within the closed environment.

Differential privacy has also been applied to training of machine learning models where the goal is to keep the  training data private while learning the parameters of the model, e.g., \cite{Duchi2014,Abadi2016a}. A successful attack using generative adversarial networks on differentialy private training mechanisms has been proposed in \cite{Hitaj2017}. Our framework differs from these models in the sense that we propose data sanitization mechanisms and privacy for user's data and not for training data.

The Pufferfish framework \cite{Kifer2014} defines privacy with respect to a set of \textit{potential secrets} and with respect to a set of assumptions about the knowledge of a potential adversary. Similarly to our framework, this framework constrains the problem to a particular secret and assumes an adversary has a limited prior knowledge. In our framework we explicitly constraint a potential adversary by defining the training data and the set of algorithms. Moreover, we also propose two techniques for data sanitization that are based on data and achieve privacy and utility within a closed environment.

A deep learning architecture for privacy-preserving mobile analytics has been proposed in \cite{AliOssia}. Their main tools is dimensionality reduction of shared features, which allows them to prevent the success of hard tasks (person identification), where a higher feature dimensionality is required. However, such methods do not work if we want to disable an easier task (e.g, gender classification).


\paragraph{Fairness.} The goal of fairness in machine learning is to ensure that predictions are not biased or discriminatory  \cite{Dwork2012}. To achieve fairness we may regularize the feature representations \cite{Zemel2013,Edwards2015,Louizos2015} or design predictors in a way that they obey the fairness constraints \cite{Hardt2016,Zafar2014}.

A fair representation with respect to a certain variable does not necessarily  imply privacy or non-identifiability of this variable. On the other hand, fairness can automatically result from the here proposed framework. Certain approaches such as the variational fair auto-encoder \cite{Louizos2015} or a censored representation  trained using an adversary \cite{Edwards2015} also achieve non-identifiability. These methods can potentially be used in our proposed framework for closed privacy. However, our focus is not on a particular method or algorithm but rather on a wider framework that may encompass various algorithms.  



\paragraph{Other.}
An orthogonal direction that offers privacy is homomorphic encryption \cite{Graepel2012}, where models are trained and tested using encrypted data, and the prediction results are also returned in an encrypted form. Homomorphic encryption has been applied to neural networks in \cite{Dowlin2016}. Such homomorphic encryption schemes are extremely computationally inefficient. Our proposed framework is as efficient as the desired utility, and again its goal is to block the entity from extracting protected information and not to block others from accessing data (encryption).


\section{Conclusions} \label{sec:conclusion}

In this work we have introduced a framework that provides privacy to users within a closed environment. We have also demonstrated its application on face image data. There are many possible extension to the presented framework. In particular, it may be interesting to  design methods that would allow sharing of users' data between various closed environments in a way consistent with privacy guarantees. Another relevant extension is to  adaptive environments where users' data is used internally to improve the models. Finally, in this work we measure utility and privacy empirically, based on a test set, which is perfectly reasonable from the practical perspective. However, we also see opportunities for a more detailed theoretical study that would lead to guarantees in a stronger statistical sense.

\clearpage

\section*{Appendix}

\paragraph{Details of SVM and RF training.}
The training procedure of the classifier works as follows: Provided data is randomly split into training and validation sets. The predictor is trained on the training set for various hyper-parameter settings and the performance on the validation set is recorded. The hyper-parameters corresponding to the best performance on the validation set are chosen and the predictor is trained on the entire dataset.

We use the linear SVM and RF implemented in \cite{scikit-learn}. The constant $C$ associated with the linear SVM is picked from the set $\{1e-3,1e-1,1,100\}$. For the RF classifiers we consider number of trees in $\{25,50,100\}$ and tree depth is set to $3,5$ or it is set  automatically by the algorithm.

\paragraph{Data Sanitization with Linear Projection.}

The sanitization function with linear projection $P$ is defined as
	\begin{IEEEeqnarray}{rCl}
		f(x) = P x \,.
	\end{IEEEeqnarray}
The linear projection $P$ is obtained by iterating the following steps, where $P$ is set to identity initially:
\begin{enumerate}
	\item The features $x$ in the training set $\sZ$ and the test set $\sZ'$ are transformed by $P$: $x = Px$.
	\item An SVM classifier is trained on the training set $\sZ$ to predict the sensitive variable $s$. SVM training details are provided in the paragraph above.
	\item A linear projection $P'$ is constructed so that the kernel of the projection corresponds to the weight vector of the trained SVM. In this way the projection ``collapses'' the dimension of the space where the training samples are linearly separable.
	\item Projection $P$ is updated to include the kernel of the projection $P'$: $P = P' P $.
\end{enumerate}
The process is repeated until we achieve a sufficiently low classification accuracy on the test set (we used 55\% in our case).

The trade-off between utility and privacy is achieved by increasing or decreasing the dimension of the kernel of the projection $P$. 

\paragraph{Data Sanitization with Maximum Mean Discrepancy Transformation.}

We extend the technique proposed in \cite{Gardner2015a}, which uses the Maximum Mean Discrepancy (MMD) statistics \cite{Gretton2007}. It is originally used for image style transfer using convolutional neural networks. 

The sanitization function is defined as 
	\begin{IEEEeqnarray}{rCl}
		f(x) = x + n(x) \,,
	\end{IEEEeqnarray}
where $n(x)$ is obtained as follows: First a dictionary $D$ is constructed from the feature vectors $x$ in the training set $\sZ$. We set $n(x)$ to be a linear combination of the features in the training set: $n(x) = D\theta(x)$ where $\theta(x)$ is the vector of coefficients. The central part of the method is the computation of the parameter vector $\theta(x)$, which we describe next.

We use the features $x$ from the training set $\sZ$ to construct two subsets: $\sX_+=\{x : (x,s,t) \in \sZ, s =1 \} $ and $\sX_-=\{x : (x,s,t) \in \sZ, s =-1 \} $, which correspond to the two classes associated with the sensitive label. Then for each user's feature vector $x$ we determine randomly the target class $s' \in \{-1,1 \}$ and define the cost function 
\begin{IEEEeqnarray}{rCl}
c(\theta(x)) = s' \times \left( \frac{1}{|\sX_+|} \sum_{x_+ \in \sX_+} k(x_+, x + D \theta(x)) - \frac{1}{|\sX_-|} \sum_{x_- \in \sX-} k(x_-, x + D \theta(x))\right)\,,
\end{IEEEeqnarray}
where $k(x, x')$ is the kernel function. We choose $k(x,x')$ to be a Gaussian kernel, $k(x,x') = \exp(- \| x - x' \|_2^2/ 2\sigma^2)$ with parameter $\sigma=0.001$ in our experiments.

The value of $\theta(x)$ is obtained by maximizing the cost function $c(\theta(x))$: 
\begin{IEEEeqnarray}{rCl}
	\max_{\theta(x)} c(\theta(x)) \,. \label{eq:MMD_sanitization}
\end{IEEEeqnarray}
The optimization problem is solved by gradient descent using step size $0.05$ and 10,000 iterations.

Intuitively, a large positive value of 
\begin{IEEEeqnarray}{rCl}
\frac{1}{|\sX_+|} \sum_{x_+ \in \sX_+} k(x_+, x + D \theta(x)) - \frac{1}{|\sX_-|} \sum_{x_- \in \sX-} k(x_-, x + D \theta(x)) \label{eq:witness}
\end{IEEEeqnarray}
indicates that $x + D(\theta)$ resembles the data in $\sX_+$, and a large negative value of \eqref{eq:witness} indicates that $x + D\theta(x)$ resembles the data in $\sX_-$. Therefore, the optimization problem in \eqref{eq:MMD_sanitization}, where $s' \in \{-1,1\}$, can be interpreted as a non-linear noise model that changes feature $x$ to be close to either $|\sX_+|$ or $|\sX_-|$.

\bibliographystyle{IEEEtran}
\bibliography{IEEEabrv,library_correct}

\begin{thebibliography}{10}
\providecommand{\url}[1]{#1}
\csname url@samestyle\endcsname
\providecommand{\newblock}{\relax}
\providecommand{\bibinfo}[2]{#2}
\providecommand{\BIBentrySTDinterwordspacing}{\spaceskip=0pt\relax}
\providecommand{\BIBentryALTinterwordstretchfactor}{4}
\providecommand{\BIBentryALTinterwordspacing}{\spaceskip=\fontdimen2\font plus
\BIBentryALTinterwordstretchfactor\fontdimen3\font minus
  \fontdimen4\font\relax}
\providecommand{\BIBforeignlanguage}[2]{{%
\expandafter\ifx\csname l@#1\endcsname\relax
\typeout{** WARNING: IEEEtran.bst: No hyphenation pattern has been}%
\typeout{** loaded for the language `#1'. Using the pattern for}%
\typeout{** the default language instead.}%
\else
\language=\csname l@#1\endcsname
\fi
#2}}
\providecommand{\BIBdecl}{\relax}
\BIBdecl

\bibitem{Narayanan2008}
A.~Narayanan and V.~Shmatikov, ``{Robust de-anonymization of large sparse
  datasets},'' \emph{IEEE Symposium on Security and Privacy}, pp. 111--125,
  2008.

\bibitem{Oh2016}
S.~J. Oh, R.~Benenson, M.~Fritz, and B.~Schiele, ``{Faceless person
  recognition; Privacy implications in social media},'' \emph{European
  Conference on Computer Vision (ECCV)}, pp. 19--35, 2016.

\bibitem{Israel2014}
S.~Israel, A.~Caspi, D.~W. Belsky, H.~Harrington, S.~Hogan, R.~Houts,
  S.~Ramrakha, S.~Sanders, R.~Poulton, and T.~E. Moffitt, ``{Credit scores,
  cardiovascular disease risk, and human capital},'' \emph{Proceedings of the
  National Academy of Sciences}, vol. 111, no.~48, pp. 17\,087--17\,092, 2014.

\bibitem{Dwork2008}
C.~Dwork, ``{Differential privacy: A survey of results},'' \emph{International
  Conference on Theory and Applications of Models of Computation}, pp. 1--19,
  2008.

\bibitem{Dwork2010}
C.~Dwork and M.~Naor, ``{On the difficulties of disclosure prevention in
  statistical databases or the case for differential privacy},'' \emph{Journal
  of Privacy and Confidentiality}, vol.~2, no.~1, pp. 93--107, 2010.

\bibitem{Kifer2011a}
D.~Kifer and A.~Machanavajjhala, ``{No free lunch in data privacy},'' \emph{ACM
  SIGMOD International Conference on Management of data}, pp. 193--204, 2011.

\bibitem{Kifer2014}
------, ``{Pufferfish: A framework for mathematical privacy definitions},''
  \emph{ACM Transactions on Database Systems}, vol.~39, no.~1, pp. 3:1--3:36,
  2014.

\bibitem{Hardt2016}
M.~Hardt, E.~Price, and N.~Srebro, ``{Equality of opportunity in supervised
  learning},'' \emph{Advances in Neural Information Processing Systems (NIPS)},
  pp. 3315--3323, Oct. 2016.

\bibitem{AliOssia}
S.~{Ali Ossia}, A.~{Shahin Shamsabadi}, A.~Taheri, H.~R. Rabiee, N.~D. Lane,
  and H.~Haddadi, ``{A hybrid deep learning architecture for privacy-preserving
  mobile analytics},'' \emph{arXiv:1703.02952}.

\bibitem{Ng2014}
H.~Ng and S.~Winkler, ``{A data-driven approach to cleaning large face
  datasets},'' \emph{IEEE International Conference on Image Processing (ICIP)},
  pp. 343--347, 2014.

\bibitem{Parkhi2015a}
O.~M. Parkhi, A.~Vedaldi, and A.~Zisserman, ``{Deep face recognition},''
  \emph{British Machine Vision Conference}, pp. 1--12, 2015.

\bibitem{Wasserman2010}
L.~Wasserman and S.~Zhou, ``{A statistical framework for differential
  privacy},'' \emph{Journal of the American Statistical Association}, vol. 105,
  no. 489, pp. 375--389, 2010.

\bibitem{Duchi2014}
M.~J. Wainwright, J.~I. Michael, and J.~C. Duchi, ``{Privacy aware learning},''
  \emph{Advances in Neural Information Processing Systems (NIPS)}, pp.
  1430--1438, 2012.

\bibitem{Abadi2016a}
M.~Abadi, A.~Chu, I.~Goodfellow, H.~B. McMahan, I.~Mironov, K.~Talwar, and
  L.~Zhang, ``{Deep learning with differential privacy},'' \emph{ACM SIGSAC
  Conference on Computer and Communications Security}, pp. 308--318, Jul. 2016.

\bibitem{Hitaj2017}
B.~Hitaj, G.~Ateniese, and F.~Perez-Cruz, ``{Deep models under the GAN:
  Information leakage from collaborative deep learning},''
  \emph{arXiv:1702.07464}, 2017.

\bibitem{Dwork2012}
C.~Dwork, M.~Hardt, T.~Pitassi, O.~Reingold, and R.~Zemel, ``{Fairness through
  awareness},'' \emph{ACM Innovations in Theoretical Computer Science
  Conference}, pp. 214--226, 2012.

\bibitem{Zemel2013}
R.~Zemel, Y.~Wu, K.~Swersky, T.~Pitassi, and C.~Dwork, ``{Learning fair
  representations},'' \emph{International Conference on Machine Learning
  (ICML)}, pp. 325--333, 2013.

\bibitem{Edwards2015}
H.~Edwards and A.~Storkey, ``{Censoring representations with an adversary},''
  \emph{International Conference on Learning Representations (ICLR)}, pp.
  1--14, Nov. 2016.

\bibitem{Louizos2015}
C.~Louizos, K.~Swersky, Y.~Li, M.~Welling, and R.~Zemel, ``{The variational
  fair autoencoder},'' \emph{International Conference on Learning
  Representations (ICLR)}, pp. 1--11, Nov. 2016.

\bibitem{Zafar2014}
M.~B. Zafar, M.~{Gomez Rodriguez}, and K.~P. Gummadi, ``{Fairness constraints :
  A mechanism for fair classification},'' \emph{arXiv:1507.05259}, 2017.

\bibitem{Graepel2012}
T.~Graepel, K.~Lauter, and M.~Naehrig, ``{ML confidential: Machine learning on
  encrypted data},'' \emph{International Conference on Information Security and
  Cryptology}, pp. 1--21, 2012.

\bibitem{Dowlin2016}
N.~Dowlin, R.~Gilad-Bachrach, K.~Laine, K.~Lauter, M.~Naehrig, and J.~Wensing,
  ``{Cryptonets: Applying neural networks to encrypted data with high
  throughput and accuracy},'' \emph{International Conference on Machine
  Learning (ICML)}, pp. 201--210, 2016.

\bibitem{scikit-learn}
F.~Pedregosa, G.~Varoquaux, A.~Gramfort, V.~Michel, B.~Thirion, O.~Grisel,
  M.~Blondel, P.~Prettenhofer, R.~Weiss, V.~Dubourg, J.~Vanderplas, A.~Passos,
  D.~Cournapeau, M.~Brucher, M.~Perrot, and E.~Duchesnay, ``{Scikit-learn:
  Machine learning in Python},'' \emph{Journal of Machine Learning Research},
  vol.~12, pp. 2825--2830, 2011.

\bibitem{Gardner2015a}
J.~R. Gardner, P.~Upchurch, M.~J. Kusner, Y.~Li, K.~Q. Weinberger, K.~Bala, and
  J.~E. Hopcroft, ``{Deep manifold traversal: Changing labels with
  convolutional features},'' \emph{arXiv:1511.06421}, 2015.

\bibitem{Gretton2007}
A.~Gretton, K.~M. Borgwardt, M.~Rasch, B.~Sch{\"{o}}lkopf, and A.~J. Smola,
  ``{A kernel method for the two-sample-problem},'' \emph{Advances in neural
  information processing systems (NIPS)}, pp. 513--520, 2007.

\end{thebibliography}

\end{document}